\documentclass[10pt, conference, compsocconf]{IEEEtran}

\usepackage{amssymb}
\usepackage[cmex10]{amsmath}

\usepackage{ifthen}
\newcommand{\lengthVersion}{short}
\newcommand{\longonly}[1]
{\ifthenelse{\equal{\lengthVersion}{long}}{#1}{}}

\newcommand{\shortonly}[1]
{\ifthenelse{\equal{\lengthVersion}{short}}{#1}{}}

\makeatletter
\newif\if@restonecol
\makeatother

\usepackage[ruled,vlined]{algorithm2e}

\usepackage{ifpdf}

\ifpdf
\PassOptionsToPackage{pdftex}{graphicx}
\fi

\usepackage{graphicx}
\ifpdf
\makeatletter
\let\orig@Gin@extensions\Gin@extensions
\def\Gin@extensions{.pdf,\orig@Gin@extensions}
\makeatother
\fi

\usepackage[caption=false,font=footnotesize]{subfig}
\usepackage{url}
\usepackage{paralist}
\usepackage[usenames]{color}
\usepackage{cite}

\newcommand{\term}[1]{\emph{#1}}
\DeclareMathOperator*{\argmax}{arg\,max}

\newcommand{\attr}[1]{\textsc{#1}} 
\DeclareMathOperator*{\Oh}{\mathrm{O}}

\newcommand{\improveFigureLayout}{
\renewcommand{\topfraction}{0.85}
\renewcommand{\textfraction}{0.1}
\renewcommand{\floatpagefraction}{0.75}
\renewcommand{\dbltopfraction}{\topfraction}
\renewcommand{\dblfloatpagefraction}{\floatpagefraction}
}
\improveFigureLayout

\newcommand{\Tab}[1]{Table~\ref{tab:#1}}
\newcommand{\Fig}[1]{Figure~\ref{fig:#1}}
\newcommand{\Alg}[1]{Algorithm~\ref{alg:#1}}

\title{COMET: A Recipe for Learning and Using Large Ensembles on Massive Data}
  \author{
    Justin D.\ Basilico, M.\ Arthur Munson, Tamara G.\ Kolda, Kevin R. Dixon, W.\ Philip Kegelmeyer\\
    Sandia National Laboratories\\
    Livermore, CA 94551, USA\\
    \{jdbasil, mamunso, tgkolda, krdixon, wpk\}@sandia.gov
  } 

\usepackage{fancyhdr}
\begin{document} 
\bstctlcite{customize_ieee_bib}
\maketitle

\begin{abstract} 
COMET is a single-pass MapReduce algorithm for learning on large-scale
data.  It builds multiple random forest ensembles on distributed
blocks of data and merges them into a mega-ensemble.  This approach is
appropriate when learning from massive-scale data that is too large to
fit on a single machine.  To get the best accuracy, IVoting should be
used instead of bagging to generate the training subset for each
decision tree in the random forest.  Experiments with two large
datasets (5GB and 50GB compressed) show that COMET compares favorably
(in both accuracy and training time) to learning on a subsample of
data using a serial algorithm.  Finally, we propose a new Gaussian
approach for lazy ensemble evaluation which dynamically decides how
many ensemble members to evaluate per data point; this can reduce
evaluation cost by 100X or more.
\end{abstract}


\section{Introduction} \label{sec:intro}

The integration of computer technology into science and daily life has
enabled the collection of massive volumes of data. However, this information cannot be
practically analyzed on a single commodity computer because the
data is too large to fit in memory.  Examples of such massive-scale data include website
transaction logs, credit card records, high-throughput biological
assay data, sensor readings, GPS locations of cell phones, etc.
Analyzing massive data requires either a) subsampling the data down to
a size small enough to be processed on a workstation; b) restricting
analysis to streaming methods that sequentially analyze fixed-sized subsets; or c) distributing the data across multiple
computers that perform the analyses in parallel.  While subsampling is
a simple solution, the models it produces are often less accurate
than those learned from all available
data~\cite{Perlich03:learning_curve, Provost99:ScalingUp}.
Streaming methods benefit from seeing all data but typically run on a
single computer, which makes processing large datasets time consuming.
Distributed approaches are attractive because they can exploit
multiple processors to construct models faster.

In this paper we propose to learn \emph{quickly} from massive volumes
of existing data using parallel computing and a divide-and-conquer
approach.  The data records are evenly partitioned across multiple compute
nodes in a cluster, and each node \emph{independently} constructs an ensemble
classifier from its data partition.  The resulting ensembles (from
all nodes) form a mega-ensemble that votes to determine classifications.
The complexities of data distribution, parallel computation, and
resource scheduling are managed by the MapReduce
framework~\cite{Dean2008:MapReduce}.  In contrast to many previous
uses of MapReduce to scale up machine
learning that require multiple passes over the data~\cite{Chu2007:MapReduce, Chang08:parallel_svm,
  Liu2010:MapReduceNeuralNet, Panda2009:PLANET,
  Gonzalez2009:parallelBP, Deodhar2010:parallelSCOAL}, this approach
requires only a \emph{single pass} (single MapReduce step)
to construct the entire ensemble.  This minimizes disk I/O and the
overhead of setting up and shutting down MapReduce jobs.

Our approach is called COMET
(short for \textbf{C}loud \textbf{O}f \textbf{M}assive
\textbf{E}nsemble \textbf{T}rees), which leverages proven learning algorithms in a novel combination.  
Each compute node constructs a random
forest~\cite{Breiman2001:RandomForests} based on its local data.
COMET employs IVoting (importance-sampled
voting)~\cite{Breiman99:PastingVotes} instead of the usual
bagging~\cite{Breiman1996:Bagging} to generate the training subsets
used for learning the decision trees in the random forest.  Chawla et
al.~\cite{Chawla04:bites} showed that IVoting produces more accurate
ensembles than bagging in distributed settings.
IVoting Random Forests combine the advantages of random forests (good
accuracy on many problems~\cite{CN2006:comparison, CKY2008:comparison}
and efficient learning with many
features~\cite{Breiman2001:RandomForests}) with IVoting's ability to
focus on more difficult examples.
The local ensembles are combined into a mega-ensemble containing
thousands of classifers in total.  Using such a large ensemble
is computationally expensive and overkill for data points
that are easy to classify. Thus, we employ a \term{lazy ensemble
  evaluation} scheme that only uses as many ensemble members as
are needed to make a confident prediction.  We propose a new
Gaussian-based approach for Lazy Ensemble Evaluation (GLEE) that is
easier to implement and more scalable than previously proposed approaches.

Our main contributions are as follows:
\begin{compactitem}
  \item \textbf{We present COMET, a novel MapReduce-based framework for distributed
    Random Forest ensemble learning.} Our method  uses a
    divide-and-conquer approach for learning on massive data and 
    requires only a single MapReduce pass for training,   unlike
    recent work using MapReduce to learn decision tree ensembles
    \cite{Panda2009:PLANET}. We also use a sampling approach called 
    IVoting rather than the usual bagging technique.
  \item \textbf{We develop a new approach for lazy ensemble evaluation based
    on a Gaussian confidence interval, called GLEE.} GLEE is easier to
    implement and asymptotically faster to compute than the
    Bayesian approach proposed by Hern\'andez-Lobato~et~al.~\cite{HernandezLobato09:InstanceBasedPruning}.
    Simulation experiments show that GLEE is as accurate as the
    Bayesian approach.
  \item \textbf{Applying COMET to two publicly available datasets (the larger of which
    contains 200M examples), 
    we demonstrate that using more data produces more accurate
    models than learning from a subsample on a single computational node.}  
    Our results also confirm that the IVoting sampling strategy significantly 
    outperforms bagging in the distributed context.
\end{compactitem}

\section{Learning on Massive Data via COMET} 
\label{sec:recipe}

COMET is a recipe for large-scale distributed ensemble learning
and efficient ensemble evaluation.  The recipe has three components:
\begin{compactenum}
  \item \textbf{MapReduce:} We write our distributed learning algorithm using
    MapReduce to easily parallelize the learning task. The mapper
    tasks build classifiers on local data partitions (``blocks'' in
    MapReduce nomenclature), and one or more reducers can combine together and output the
    classifiers. The learning phase only takes a single MapReduce job. 
    If the learned ensemble is large and/or the number of data
    points to be evaluated is large, evaluation can also be
    parallelized using at most two MapReduce jobs.
  \item \textbf{IVoting Random Forest:} Each mapper builds an
    ensemble based on its local block of data (assigned by MapReduce). The mapper runs a variant of
    random forests that replaces bagging with IVoting (described in \ref{sec:ivoting}). 
    IVoting has the advantage that it gives more weight to difficult examples.
    Unlike boosting \cite{Freund1996:boosting}, however, each model in the ensemble votes with equal
    weight, allowing us to trivially merge the ensembles from all
    mappers into a single large ensemble.
  \item \textbf{Lazy Ensemble Evaluation:} 
    Many inputs are ``easy'' and the vast majority of the ensemble members
    agree on the classification.  For these cases, querying a small sample
    of the members is sufficient to determine the ensemble's prediction
    with high confidence.  Lazy ensemble evaluation significantly lowers the
    prediction time for ensembles.
\end{compactenum}
The rest of this section describes these three components in more
detail.

\subsection{Exploiting MapReduce for Distributed Learning}
\label{sec:mapreduce}

We take a coarse-grained approach to distributed
learning that minimizes communication and coordination between compute
nodes.  
We assume that the training data is partitioned randomly into blocks 
in such a way that class distributions are roughly the same across all blocks.
 Such shuffling can be accomplished in a simple
pre-processing step that maps each data item to a random block.

In the learning phase, each mapper independently learns a predictive model from
an assigned data block.  The learned models are aggregated together into a
final ensemble model by the reducer. This is the only step that requires
internode communication, and only the final models are transmitted (not
the data).  Thus, \emph{we only require a single MapReduce pass for training}. 

We implement the above strategy in the MapReduce framework
\cite{Dean2008:MapReduce} because the framework's abstractions match
our needs, although other parallel computing frameworks (e.g., MPI)
could also be used.  To use MapReduce, one loads the input data into
the framework's distributed file system and defines map and reduce
functions to process key-value pair data during Map and Reduce stages,
respectively.  Mappers execute a map
function on an assigned data block (usually read from the node's
local file system).  The map function produces zero or more key-value
pairs for each input; in our case, the values correspond to
learned trees (with random keys).  During the Reduce stage, all the pairs emitted
during the Map stage are grouped by key and passed to reducer
nodes that run the reduce function.  The reduce function receives one
key and all the associated values produced by the Map stage.  Like the
map function, the reduce function can emit any number of key-value
pairs.  Resulting pairs are written to the distributed file system.
The MapReduce framework manages data partitioning, task scheduling,
data replication, and restarting from failures. The reducer(s) write the learned trees to one or more output files.

\longonly{
\begin{algorithm} 
\caption{Learning Phase: Map Step}
\label{alg:map}
\small
\KwIn{$D$, a data block consisting of training examples}
\KwIn{$p$, number of ensemble partitions to create}
\KwOut{$E$, a classifier model}
$D = $data from assigned block; \tcp{load block}
$E = \mathrm{IVote}(D)$; \tcp{learn ensemble from data block}
\ForEach{$T \in E$}
{$r = $ uniform random number from $\{1,2,\ldots,p\}$\;
\textbf{emit} $\langle r, T \rangle$ \tcp{emit key-value pair}
}
\end{algorithm} 
}

The map and reduce functions for distributed ensemble learning are
straightforward.  The map function \longonly{is shown in \Alg{map}; it
}trains an ensemble on its local data block and then emits the learned
trees.  Each tree is emitted with a random key 
to automatically partition the ensemble
across the reducers.
\longonly{The reduce function is shown in \Alg{reduce} and combines
the trees for the key assigned to it. Thus, if trees are emitted to a
single partition ($p=1$), all trees will be reduced to one output
file.  If $p > 1$, each reducer will receive approximately $1/p$ of
the randomly assigned trees, and there will be $p$ output files.}

\longonly{
\begin{algorithm}
\caption{Learning Phase: Reduce Step}
\label{alg:reduce}
\small 
\KwIn{Key $k$ and associated list of trees $\mathcal{T}_k$.}
\KwOut{Ensemble $E$ with trees from all mappers, or random subset of full ensemble if keys emitted by Map Step not constant.}
\tcp{Merge ensembles learned by different mappers}
$E = \emptyset$\;
\ForEach{$T \in \mathcal{T}_k$}
{
$E = E \cup T$\;
}
{\textbf{emit} $\langle \texttt{reducer\_id}, E \rangle$}
\end{algorithm}
}

\subsection{IVoting Random Forests for  Mega-Ensembles}
\label{sec:ivoting}

Each mapper \longonly{\Alg{map} }in COMET builds an ensemble from the local
data partition using IVoting.  IVoting (Importance-sampled Voting)
\cite{Breiman99:PastingVotes} builds an ensemble by repeatedly
applying the base learning algorithm (e.g., decision tree induction
\cite{cartbook,Quinlan1986:Induction}) to small samples called
\term{bites}.  Unlike bagging \cite{Breiman1996:Bagging}, examples are
sampled with non-uniform probability.  Suppose that $k$ IVoting
iterations have been run, producing ensemble $E_k$ comprised of $k$
base classifiers.  To form the $k+1^{\rm st}$ bite, training examples
$(x,y)$ are drawn randomly.  If $E_k$ incorrectly classifies $x$,
$(x,y)$ is added to training set $B_{k+1}$.  Otherwise $(x,y)$ is
added to $B_{k+1}$ with probability $e(k)/(1-e(k))$, where $e(k)$ is
the error rate of $E_k$.  This process is repeated until $|B_{k+1}|$
reaches the specified bite size $b$; $b$ is typically smaller than the size of the full data.  Out-of-bag (OOB)
\cite{Breiman96:oob} predictions are used to get unbiased estimates of
$e(k)$ and $E_k$'s accuracy on sampled points $x$.  The OOB prediction
for $x$ is made by voting only the ensemble members that did not see
$x$ during training, i.e., $x$ was outside the base models' training
sets.

IVoting's sequential and weighted sampling is reminiscent of boosting
\cite{Freund1996:boosting} and is similar to boosting in
terms of accuracy \cite{Breiman99:PastingVotes}.  IVoting differs from
boosting in that each base model receives equal weight for
deciding the ensemble's prediction.  This property simplifies merging
the multiple ensembles produced by independent IVoting runs.

Breiman \cite{Breiman99:PastingVotes} showed that IVoting sampling
generates bites containing roughly half correct and half incorrect
examples.  Our implementation (\Alg{ivote}) draws, with replacement,
50\% of the bite from the examples $E_k$ correctly classifies and 50\%
from the examples $E_k$ incorrectly classifies (based on OOB
predictions).  This implementation avoids the possibility of drawing
and rejecting large numbers of correct examples for ensembles with
very high accuracy.

\begin{algorithm}
\caption{IVoting --- Ensemble learning that samples correct \& incorrect examples in equal proportions.}
\label{alg:ivote}
\small
\KwIn{Dataset $D \in (\mathcal{X},\mathcal{Y})^{*}$; Ensemble size $m$; Bite size $b \in \mathbb{N}$; Base learner $L : (\mathcal{X}, \mathcal{Y})^{*} \rightarrow ( \mathcal{X} \rightarrow \mathcal{Y}) $}
\KwOut{Ensemble $E$}
Initialize $D^{+}_{0} = D$, $D^{-}_{0} = D$, $V_{oob}[\cdot, \cdot] = 0$, $E = \emptyset$\;
\For{$i \in [1, m]$}
{
   \tcp{Create the bite to train on.}
    $B^{+}_{i}$ = $b/2$ uniform random samples from $D^{+}_{i-1}$\;
    $B^{-}_{i}$ = $b/2$ uniform random samples from $D^{-}_{i-1}$\;
    $B_i = B^{+}_{i} + B^{-}_{i}$\;
    \tcp{Train a new ensemble member.}
    $T_{i} = L(B_i)$\;
    Add $T_i$ to $E$\;
    \tcp{Update running votes.}
    \For{$(x_j, y_j) \notin B_i$}
    {
        $V_{oob}[j, T_i(x_j)]$ += 1\;
    }
    $D^{+}_{i} =  \{(x_j, y_j) \in D \mid y_j = \text{arg max}_{z} V_{oob}[j, z] \}$\;
    $D^{-}_{i} =  \{(x_j, y_j) \in D \mid y_j\neq \text{arg max}_{z} V_{oob}[j, z] \}$\;
} 
\Return{$E$}\;
\end{algorithm} 

Any classification learning algorithm could be used for the base learner
in IVoting.  Our experiments use decision trees
\cite{Quinlan1986:Induction,Quinlan1993:C45} because they generally
form accurate ensembles \cite{Bauer99:voting}.  The trees are grown to
full size (i.e., each leaf is pure or contains fewer than ten training
examples) using information gain as the splitting criterion.  We use
full-sized trees because they generally yield slightly more accurate
ensembles \cite{Bauer99:voting}.  To increase the diversity of trees
and reduce training time for data sets with large numbers of features,
only a random subset of features are considered when choosing the test
predicate for each tree node.  This attribute subsampling is used in
random forests and has been shown to improve performance and decrease
training time \cite{Breiman2001:RandomForests}.  We employ the random
forest heuristic for choosing the attribute sample size 
\longonly{$d'$:
\begin{align*}
d' &= \lfloor 1 + \log_2 d \rfloor
\end{align*}
}
\shortonly{$d' = \lfloor 1 + \log_2 d \rfloor$,}
where $d$ is the total number of attributes.

\subsection{Lazy Ensemble Evaluation via a Gaussian Approach}
\label{sec:glee}

A major drawback to large ensembles is the \longonly{runtime} cost of
querying all ensemble members for their predictions.  In practice,
many data points are easy to classify: the vast majority of the
ensemble members agree on the classification.  For these cases,
querying a small sample of the members is sufficient to determine the
ensemble's prediction with high confidence.

We exploit this phenomena via lazy ensemble evaluation. \term{Lazy
  ensemble evaluation} is the strategy of only evaluating as many
ensemble members as needed to make a good prediction on \emph{a case
  by case basis for each data point}.  Ensemble voting is stopped when
the ``lazy'' prediction has high probability of being the same as the
prediction from the entire ensemble.
 The risk that lazy evaluation stops voting too early (i.e., the
probability that the early prediction is different from what the full
ensemble prediction would have been) is bounded by a user-specified
parameter $\alpha$. \Alg{lazy} lists the lazy ensemble evaluation
 procedure.  Let $x$ be a data point to classify using
ensemble $E$, with $E$ containing $m$ base models.  Initially all $m$
models are in the unqueried set $U$.  In each step, a model $T$ is
randomly chosen and removed from $U$ to vote on $x$; the vote is added to
the running tallies of how many votes each class has received.  Based
on the accumulated tallies and how many ensemble members have not yet
voted, the stopping criterion decides if it is safe to stop and return the
classification receiving the most votes.  If it is not safe, a new
ensemble member is drawn, and the process is repeated until it is safe
to stop or all $m$ ensemble members have been queried.  
Note that lazy evaluation is agnostic to whether the base models are
correlated.  Its goal is to approximate the (unmeasured) vote
distribution from a sample of votes, and the details of the process
generating the votes are irrelevant.

\begin{algorithm} 
\caption{Lazy Ensemble Evaluation}
\label{alg:lazy}
\small
\KwIn{Input $x \in \mathcal{X}$}
\KwIn{Ensemble $E$ with $m$ members of $f : \mathcal{X} \rightarrow \{1, ..., c\}$}
\KwIn{$\alpha$, max.\ disagreement freq.\ for lazy vs.\ full eval.}
\KwIn{Vote stopping criteria $\mathrm{Stop} : (\mathbb{N}_0^c, \mathbb{N}_1, \mathbb{R} \in [0,1]) \rightarrow \{\mathrm{true}, \mathrm{false}\}$}
\KwOut{Approximate prediction from $E$ for input $x$.}
Set $U = E$, $V = [0, ..., 0]$, $|V| =c$\;

\For{$i \in [1, m]$}
{
    Sample $T$ uniformly from $U$\;
    Remove $T$ from $U$\;
    Evaluate $v_i = T(x)$\;
    Increment $V[v_i]$\;

    \If{$\mathrm{Stop}(V, m, \alpha)$}
    {
        \Return{$\argmax_{i} V[i]$}\;
    }

} 
\Return{$\argmax_{i} V[i]$}
\end{algorithm} 

\longonly{
There are many different ways to bound the probability that a current
classification estimate $\hat{o}$ is different from the true
ensemble's classification $o$ (found by voting all ensemble members),
including the Chebyshev Inequality, Chernoff Inequality, and fitting a
parametric distribution model to the votes observed thus far.  
}

In binary categorization, the vote of each base model can be modeled
as a Bernoulli random variable.  Accordingly, the distribution of
votes for the full ensemble is a binomial distribution with
proportion parameter $p$.  Provided that the number of members queried
\longonly{by the $i^{\rm th}$ step in \Alg{lazy}} is sufficiently large,
we can invoke the Central Limit Theorem and approximate the binomial
distribution with a Gaussian distribution.

We propose Gaussian Lazy Ensemble Evaluation (GLEE), which uses the
Gaussian distribution to infer a $(1-\alpha)$ confidence interval
around the observed mean $\hat{p}$.  The interval is used to test the
hypothesis that the unobserved proportion of positive votes $p$ falls
on the same side of 0.5 as $\hat{p}$ (and consequently, 
that the current estimated classification agrees with the full ensemble's classification).
If 0.5 falls outside the interval, GLEE
rejects the null hypothesis that $p$ and $\hat{p}$ are on different
sides of 0.5 and terminates voting early.  Formally, denote the
interval bounds as $\hat{p} \pm \rho\delta$, where
\begin{align*}
  \delta &= z_{\alpha/2} \frac{\sigma}{\sqrt{n}} = z_{\alpha/2} \frac{\sqrt{\hat{p} (1-\hat{p})}}{\sqrt{n}}
\end{align*}
and
\begin{align*}
  \rho &= \begin{cases}
    \sqrt{\frac{m-n}{m-1}} & \text{ if } n > 0.05 m \\
    1  & \text{ otherwise.}
\end{cases}
\end{align*}
The critical value $z_{\alpha/2}$ is the usual value from the standard
normal distribution.  The finite population correction (FPC) $\rho$
accounts for the fact that base models are drawn from a finite
ensemble.  Intuitively, uncertainty about $p$ shrinks
as the set $U$ becomes small.  To ensure the Gaussian approximation is
reasonable, GLEE only stops evaluation only once some minimum number
of models have voted.  Using simulation experiments we found 15, 30,
and 45 reasonable for $\alpha \ge 10^{-2}$, $\alpha = 10^{-3}$, and
$\alpha = 10^{-4}$, respectively.
 (Simulation methodology is described in
Section~\ref{sec:evt-comparison}.)

The above hypothesis test only requires the lower bound (if $\hat{p} >
0.5$) or the upper bound (if $\hat{p} < 0.5$).  Consequently we can
improve GLEE's statistical power by computing a one-sided interval;
i.e., use $z_{\alpha}$ instead of $z_{\alpha/2}$.  When the GLEE
stopping criteria is invoked, the \term{leading class} (the class with
the most votes so far) is treated as class 1, and the \term{runner-up
  class} is treated as class 0.\footnote{This class relabeling trick
  also enables direct application of GLEE to multiclass problems.
  \longonly{The hypothesis test is then between the class with the
    most votes and the second most votes.}} GLEE stops evaluation
early if the lower bound $\hat{p} - \delta$ is greater than
0.5.\footnote{This one-sided test is slightly biased because the
  procedure effectively chooses to compute a lower or upper bound
  after ``peeking'' at the data to determine which class is the
  current majority class.  The results in Section~\ref{sec:evt-comparison}
  show that the relative error of the lazy prediction is bounded by
  $\alpha$ despite this bias.}

Hern\'{a}ndez-Lobato et
al.~\cite{HernandezLobato09:InstanceBasedPruning} present another way
of deciding when to stop early using Bayesian inference.  We compare
to this method in Section~\ref{sec:evt-comparison} and refer to it as
{Madrid Lazy Ensemble Evaluation} (MLEE).  In MLEE, the distribution
of vote frequencies for different classes is modeled as a multinomial
distribution with a uniform Dirichlet prior.  The posterior
distribution of the class vote proportions is updated at each
evaluation step to reflect the observed base model prediction.  MLEE
computes the probability that the final ensemble predicts class $c$ by
combinatorially enumerating the possible prediction sequences for the
as-yet unqueried ensemble members, based on the current posterior
distribution.  Like GLEE, ensemble evaluation stops when the
probability of some class exceeds the specified confidence level or
when all base models have voted. MLEE is exponential in the number of
classes but is $\Oh(m^2)$ for binary classification ($m$ ensemble
members), and approximations exist to make it tractable for some
multi-class problems
\cite{MartinezMunoz2009:InstanceBasedMultiClassPruning}.

\subsection{Lazy Evaluation in a Distributed Context}
\label{sec:committee-eval}

Large ensembles (too large to fit into memory) can make predictions on
massive data (also too large to fit into memory) using \term{lazy
  committee evaluation}.  Each input is first evaluated by a
sub-committee---a random subset of the ensemble small enough to fit in
memory---using a lazy evaluation rule.  In most cases, the
sub-committee will be able to determine the ensemble's output with
high confidence and output a decision.  In the rare cases where the
sub-committee cannot confidently classify the input, the input is sent
to the full ensemble for evaluation.

Lazy committee evaluation requires two MapReduce jobs.  In the first
job each mapper randomly chooses and reads one of the $p$ ensemble
partitions to be the local sub-committee and lazily evaluates the
sub-committee on the mapped test data it receives; only one test input
needs to be in memory at a time.  If the sub-committee reaches a
decision, the input's identifier and label are written directly to the
file system.  Otherwise, a copy of the input is written to each of $p$
reducers by using keys $1,2,\ldots,p$.  In the reduce stage, each
reducer reads a different ensemble partition so that every base model
votes exactly once.  Reducers output the vote tallies for each input
they read, keyed on the input identifier.  The second job performs an
identity map with reducers that sum the vote tallies for each input;
reducers output the class with the most votes keyed to the input's
identifier.  Combining the two sets of label outputs, from the first
map and second reduce, provides a label for every input.

\section{Comparison of Lazy Evaluation Rules}
\label{sec:evt-comparison}

\begin{figure*}[tbhp]
  \centering
    \subfloat[Average savings from lazy evaluation when requiring 99.99\% confidence that lazy prediction matches full ensemble prediction.]{\label{fig:simulation-vary-ensemble-size}\includegraphics[scale=0.7]{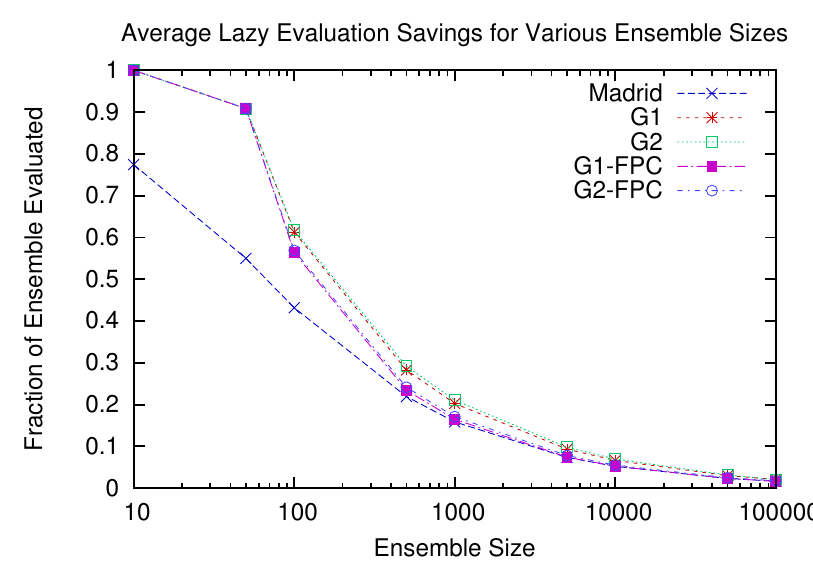}}~~
    \subfloat[Average evaluation savings for ensemble with 10K members at different confidence levels ($1-\alpha$).]{\label{fig:simulation-vary-alpha-rel}\includegraphics[scale=0.7]{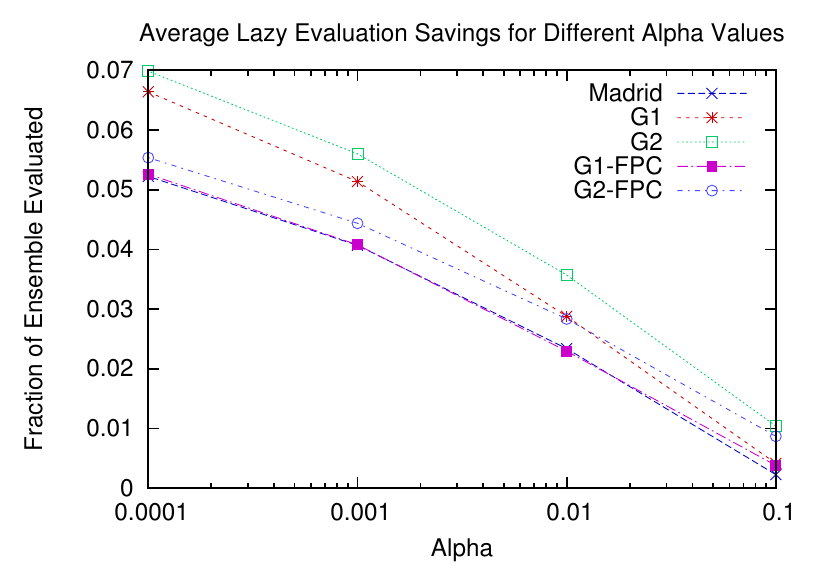}}~~
    \subfloat[Relative error of all five lazy evaluation rules is less than the specified error bound (that is, $\alpha$) for an ensemble with 10K members.]{\label{fig:simulation-vary-alpha-err}\includegraphics[scale=0.7]{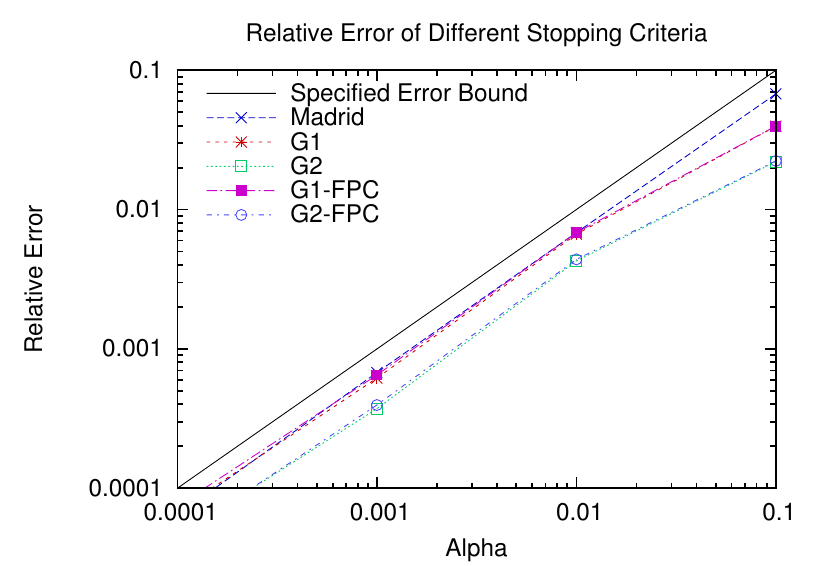}}
    \caption{Lazy ensemble evaluation drastically reduces the cost of
      evaluating large ensembles and introduces predictable, small
      relative error.  Graphs show five variants of lazy evaluation
      rules applied to simulated ensembles (see text).  The rules are
      Madrid (MLEE), G1 (Gaussian one-tail), G2 (Gaussian two-tail),
      G1-FPC (G1 with finite population correction) and G2-FPC (G2
      with finite population correction).  All five methods provide
      similar speed-ups for ensembles with 500 members or more, with
      G1-FPC and Madrid requiring slightly fewer votes than the
      others.  The relative errors of G1-FPC and Madrid rules are
      nearly identical. }
  \label{fig:simulation}
\end{figure*}

This section explores the efficacy of the GLEE rule across a wide
range of ensemble sizes and for varying confidence levels.  We
simulate votes from large ensembles to explore the rule's behavior and
to compare it to the MLEE rule.  

The stopping thresholds for both methods are pre-computed and stored
in a table that is indexed by the number of votes received by the
leading class.  One table is needed per ensemble size $m$.
Pre-computing and caching the thresholds is necessary to make MLEE
practical for large ensembles.  Once the thresholds are computed,
evaluating them requires an array lookup and comparison.  

Computing the large factorials in MLEE requires care to avoid
numerical overflow.  Mart\'{\i}nez-Mu\~{n}oz et
al.~\cite{MartinezMunoz2009:InstanceBasedMultiClassPruning} suggest
representing numbers in their prime factor decomposition to avoid
overflow; this approach requires $\Oh(m^3/\log m)$ time to compute
the table.\footnote{By the Prime Number Theorem, there are
  $\Oh(m/\log m)$ prime numbers less than $m$.  Thus, operations on
  numbers represented by prime factors take $\Oh(m/\log m)$ time.
  These operations are inside two nested $\Oh(m)$ loops.}  We instead
compute the factorials for MLEE in log-space which produces the same
results and requires $\Oh(m^2)$ total complexity.  In comparison,
computing the threshold table for GLEE takes $\Oh(m)$ time.  This
difference is significant for very large ensembles
(Table~\ref{tab:cache-time-summary}).

\begin{table}[htbp]
\caption{Time to Pre-compute Stopping Thresholds
}
\label{tab:cache-time-summary} 
\centering
\begin{tabular}{lcccccc}
        & \multicolumn{6}{c}{Ensemble Size} \\
	& 100	& 1K		& 10K	& 100K	& 200K	& 1M \\
\hline
GLEE		& 1ms	& 1ms	& 3ms	& 12ms	& 17ms	& 57ms \\
MLEE		& 4ms	& 38ms	& 2.35s	& 2.76m	& 10.12m	& 3.70h \\
\end{tabular}
\vspace{-3ex}
\end{table}

Ensemble votes are simulated as follows.  A uniform random number $p
\in [0,1]$ is generated to be the proportion of ensemble members that
vote for class 1.  The correct label for the example is 1 if $p \geq
0.5$ and 0 otherwise.  Each model in the ensemble votes by sampling
from a Bernoulli random variable with probability $\Pr(x = 1) = p$.
The ensemble is evaluated until the stopping criterion is satisfied or
all $m$ ensemble members have voted. The lazy predictions under the
different stopping rules and the prediction from evaluating the full
ensemble are compared to the correct label to determine their relative
accuracies.  This process is repeated 1 million times to simulate making
predictions for 1 million data points.  

We report the results in terms of the average fraction of the ensemble
evaluated before lazy evaluation stopped and \term{relative error}.
\term{Relative error} is the relative increase in error rate from lazy
evaluation (i.e., 1 - $\frac{\text{lazy accuracy}}{\text{full
    accuracy}}$).

\Fig{simulation-vary-ensemble-size} compares five approaches to lazy
ensemble evaluation.
The five different approaches are Madrid (MLEE),
G1 (Gaussian one-tail), G2 (Gaussian two-tail), G1-FPC (G1 with finite
population correction) and G2-FPC (G2 with finite population
correction).
All five methods provide similar speed-ups for $m \ge 500$, with
G1-FPC and Madrid requiring slightly fewer votes than the others.
More importantly, we see that there is a significant benefit to
applying these methods for ensembles with as few as 100 members and
that the benefit becomes greater as the ensemble size grows.

In Figures \ref{fig:simulation-vary-alpha-rel} and
\ref{fig:simulation-vary-alpha-err}, we fix the ensemble size at
$m=10000$ and vary $\alpha$.  As one might expect, more stringent
values of $\alpha$ (i.e., smaller) require evaluating more of the
ensemble. \Fig{simulation-vary-alpha-err} shows that $\alpha$ upper
bounds the relative error for all methods.  For example, with
$\alpha=0.01$ the relative error is less than 1\%, and less than 3\%
of the ensemble needs to be evaluated by G1-FPC and MLEE.
In the rest of the paper we use G1-FPC for lazy evaluation and will
refer to it as the GLEE rule.  Section~\ref{sec:experiments} presents
results for GLEE on real data.

\section{Experiments} \label{sec:experiments}

To understand how well our COMET approach
performs we ran a set of experiments on two large
real-world datasets.

\subsection{Datasets}

The data sets are described in detail below; the characteristics are summarized in \Tab{data-summary}.

\subsubsection{ClueWeb09 Dataset}

ClueWeb09 \cite{Callan2009:ClueWeb} is a web crawl of over 1~billion
web pages (approximately 5TB compressed, 25TB uncompressed).  For this
dataset we use language categorization as the prediction task.
Specifically, the task is to predict if a given web page's language is
English or non-English.
The features are proportions of alpha-numeric characters ($0-9$, $a-z$, $A-Z$) plus one additional feature for any other character, for a total of 63 features.

We used MapReduce to extract features for each web page and randomly divide the data into blocks by mapping each example to a random key.
Preprocessing the full ClueWeb dataset took approximately 2 hours on our Hadoop cluster and created 1000 binary files totaling approximately 259 GB and containing nearly 1B examples.
From this, we randomly extracted 200M training and 1M testing examples.
The training data was divided into 200 blocks, each approximately 1/4GB in size and containing 1M examples.

\begin{table}[tbp]
\caption{Dataset Characteristics}
\label{tab:data-summary} 
\centering
\begin{tabular}{lcccc}
\textsc{Name} & \textsc{Train} & \textsc{Test} & \textsc{Features} & \textsc{\% Positive} \\
\hline
ClueWeb & 200M & 1M &  63 & 48.4\% \\
eBird   & 1M & 400K & 1143 & 31.8\% \\
\end{tabular}
\vspace{-3ex}
\end{table}

\subsubsection{eBird}

The second dataset we use to evaluate COMET is the US48 eBird
reference dataset \cite{ebirddata2010}.  Each record corresponds to a
checklist collected by a bird watcher and contains counts of how many
birds, broken down by species, were observed at a given location and
time.  In addition to the count data, each record includes attributes
describing the environment in which the checklist was collected (e.g.,
climate, land cover), the time of year, and how much effort the
observer spent.  The eBird data tests how well COMET scales for
problems with data having hundreds of attributes.

The prediction task in our experiment is to predict if an American
Goldfinch (Carduelis tristis) will be observed at a given place and
time based on the environmental and data collection attributes.  We
chose American Goldfinches because they are widespread throughout the
United States (and thus, frequently observed) and exhibit complex
migration patterns that vary from one region to another (making the
prediction task hard).  We used the data from 1970--2008 for training
and the data from 2009 for testing.  All non-zero counts were
converted to 1 to create a binary prediction task.  
We used all
attributes except meta-data attributes intended for data filtering
(\attr{country}, \attr{state\_province}, \attr{sampling\_event\_id},
\attr{latitude}, \attr{longitude}, \attr{observer\_id},
\attr{subnational2\_code}).  
\longonly{See Appendix~\ref{app:preproc} for full pre-processing details.}

After pre-processing, the data set contains 1.4M examples and requires 5GB of compressed storage. 
We subdivided the data into 14 training and 6 testing blocks.
Each block contains 70K examples and requires 1/4 GB of storage.

\subsection{Implementation Details}

For our experiments, we used Hadoop (version 0.21), which includes MapReduce and
the Hadoop distributed file system  (HDFS).
We used the machine learning algorithm implementations from the open-source Cognitive Foundry \cite{Basilico2008:Foundry}.

All experiments were run on a
cluster with 65 worker nodes. Each worker node has one quad-core Intel
i-720 (2.66 Ghz) processor, 12 GB of memory, four 2 TB disk drives, and 1Gb
Ethernet networking. 
Each worker node was configured to execute up to four map or reduce tasks concurrently. 
To make running times directly comparable, we ran the serial algorithm on a worker
node with a copy of the training data sample on the local
file system. 

We loaded the data into HDFS with a big enough block size to ensure each file was contained in one block (i.e., 256MB, vs.\ the default 64MB block size).
Large block sizes improve accuracy by allowing IVoting to sample from more diverse examples\longonly{, at the expense of spending more time per worker node}.

\begin{figure*}
  \centering
  \subfloat[Accuracy Comparison]{\label{fig:clueweb-acc}\includegraphics[scale=0.7]{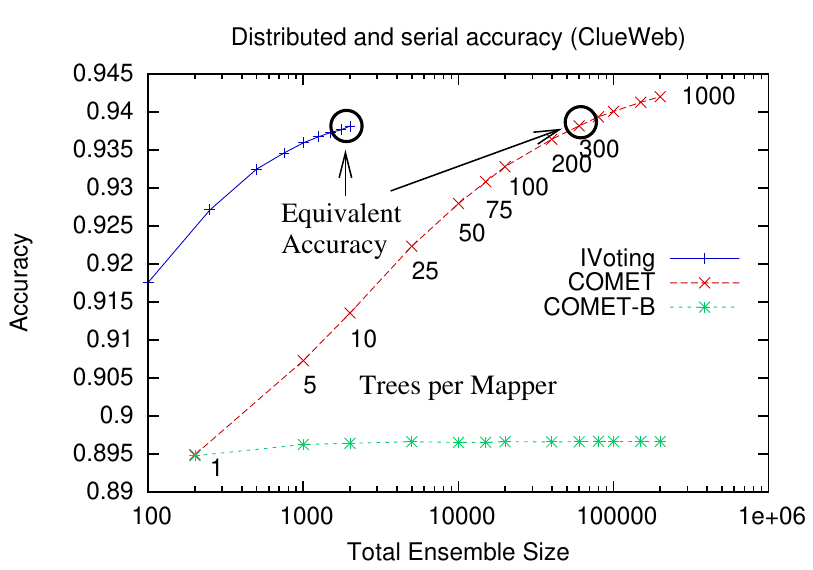}}~~
  \subfloat[Training Time Comparison]{\label{fig:clueweb-time}\includegraphics[scale=0.7]{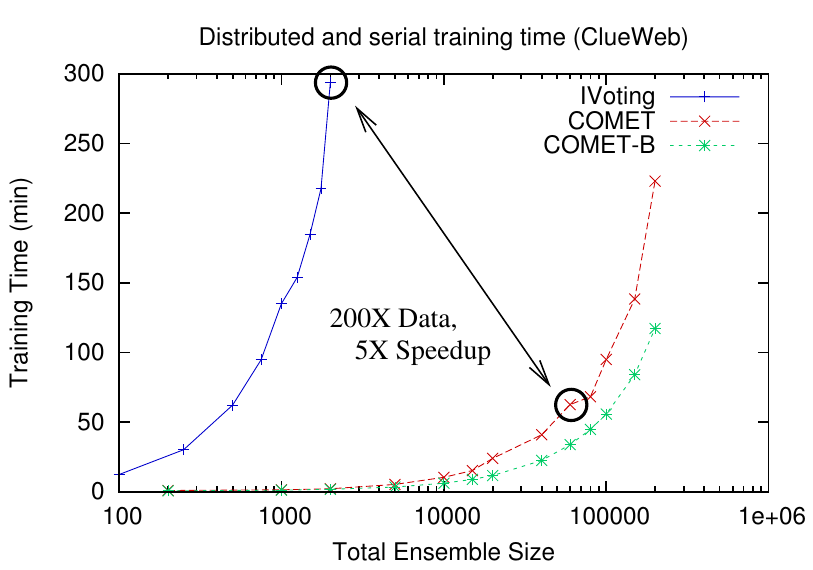}}~~
  \subfloat[Vary Training Data \& Ensemble Size]{\label{fig:clueweb-data}\includegraphics[scale=0.7]{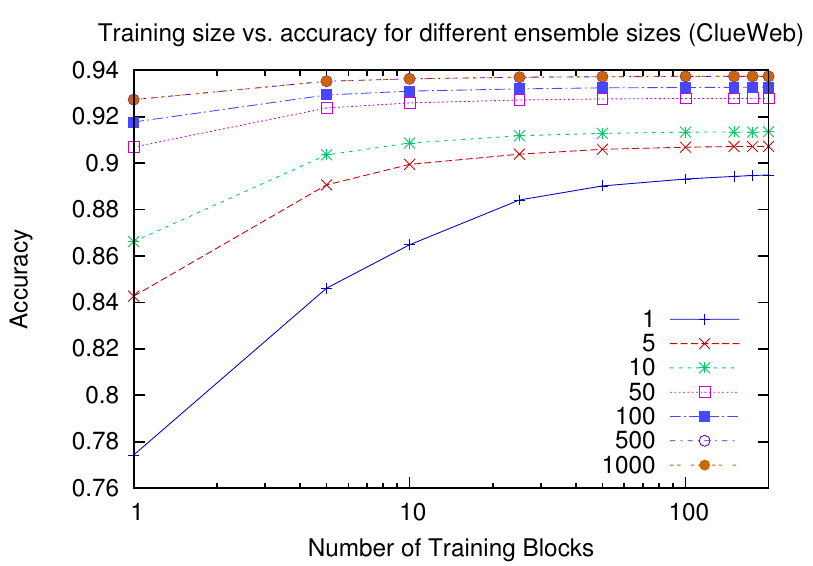}}
  \caption{On ClueWeb data, COMET can achieve better accuracy in less
    time than IVoting run serially on a subsample (IVoting
    line)---even though the training data size is 200M examples for
    COMET (distributed in 200 blocks) vs.\ 1M for serial IVoting.
    Circles denote equal accuracy.  For comparison,
    Figures~\subref{fig:clueweb-acc} and~\subref{fig:clueweb-time}
    also plot the performance of a COMET variant that uses distributed bagging
    instead of distributed IVoting (COMET-B line).
    Figure~\subref{fig:clueweb-data} illustrates varying the number of
    training data blocks (1M examples per block).  Different lines
    correspond to varying size of local ensemble.  Lines for ensemble
    sizes 500 and 1000 are superimposed. Accuracy plateaus at
    approximately 40 blocks.}
  \label{fig:clueweb-results}
  \vspace{-3ex}
\end{figure*}

\begin{figure*}
  \centering
  \subfloat[Accuracy Comparison]{\label{fig:ebird-acc}\includegraphics[scale=0.7]{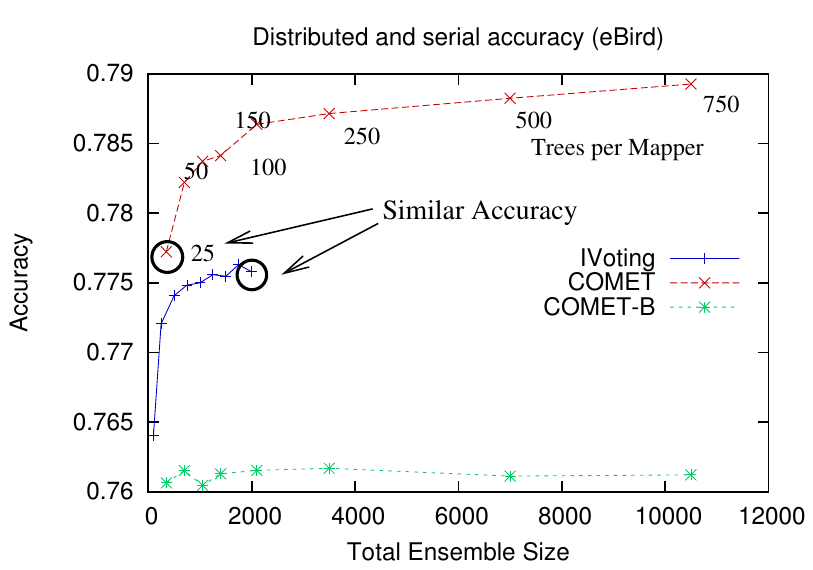}}~~
  \subfloat[Training Time Comparison]{\label{fig:ebird-time}\includegraphics[scale=0.7]{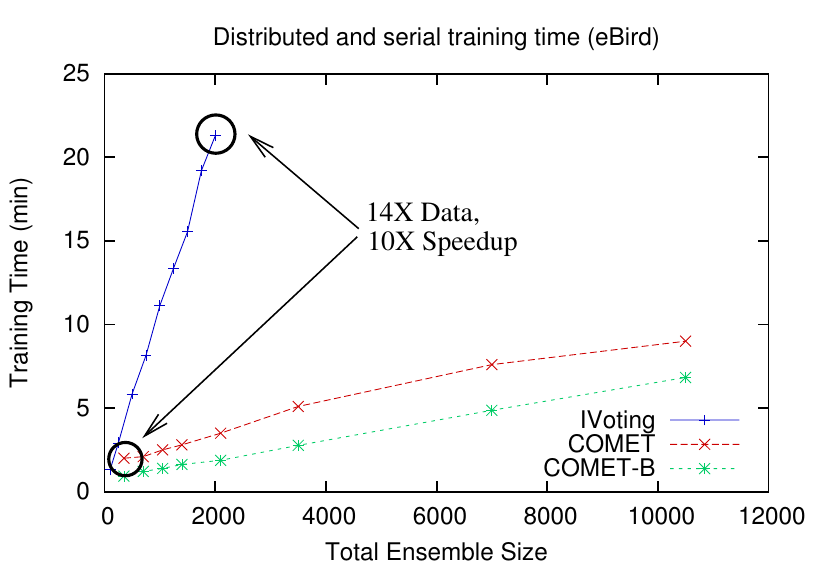}}~~
  \subfloat[Vary Training Data]{\label{fig:ebird-data}\includegraphics[scale=0.7]{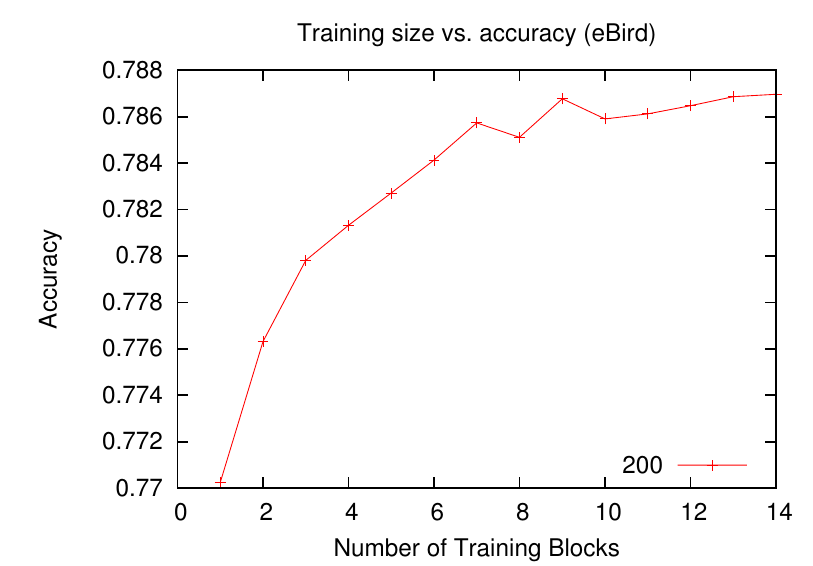}}
  \caption{Using 1M eBird training examples (distributed in 14
    blocks), COMET is more accurate in less time than IVoting applied
    serially to a 70K subsample (IVoting line)---despite processing
    14X more data.  Circles denote similar accuracies.  The COMET-B
    line in Figures~\subref{fig:ebird-acc} and~\subref{fig:ebird-time}
    shows the performance of COMET if distributed bagging is used
    instead of distributed IVoting. Figure~\subref{fig:ebird-data}
    illustrates varying the number of training data blocks (70K
    examples per block) with 200 ensemble members trained per block.}
  \label{fig:ebird-results}
  \vspace{-3ex}
\end{figure*}

In all experiments the bite size $b$ was set to 100K for ClueWeb and
10K for eBird.  These values were chosen by running IVoting for 1000 iterations on one
data block with different bite sizes and measuring the accuracy on the
test data.  For eBird, accuracy peaked at 10K (\Tab{bitesize}),
possibly because larger bite sizes reduced the diversity of the base
models.  For ClueWeb, accuracy started to plateau around 100K
(\Tab{bitesize}).  While larger bite sizes yielded small improvements,
they also resulted in trees with big enough memory footprints to
significantly limit how many ensemble members could be trained per
core.

\begin{table}
  \caption{\label{tab:bitesize}Accuracy for Different Bite Sizes. We used the bite size that corresponded to a leveling off of improvement, i.e., 100K for ClueWeb and 10K for eBird.}
  \centering
  \begin{tabular}{ccc}
                       & \textsc{ClueWeb}  & \textsc{eBird} \\
    \textsc{Bite Size} & \textsc{Accuracy} & \textsc{Accuracy} \\
    \hline
100   &   n/a     &   0.7265 \\
500   &   n/a     &   0.7496 \\
1K    &   0.8911  &   0.7614 \\
5K    &   0.9089  &   0.7753 \\
10K   &   0.9163  &   0.7755 \\
50K   &   0.9316  &   0.7713 \\
100K*  &  0.9359  &   0.7699 \\
150K	&   0.9370  &   n/a \\
200K	&   0.9377  &   n/a \\
\hline
  \end{tabular}\\
  * eBird bite size was 70K (approx.\ data partition size).
  \vspace{-3ex}
\end{table}

In GLEE, the straightforward way to sample models (without replacement) from the ensemble is to generate a new random number for each ensemble member that is evaluated. 
If the cost of generating a random number is relatively expensive, lazy evaluation may not provide enough of a speed-up and may even slow down ensemble evaluation.
To avoid this, our GLEE implementation permutes the ensemble order once at load time.
Each ensemble evaluation is started from a different random index in this order.
Thus, only a single random number is generated per ensemble prediction.

\subsection{Results}

We first compare COMET to subsampling (i.e., IVoting Random Forests
run serially on a single block of data) to measure the benefits of
learning from all data.  Accuracies are computed using full ensemble
evaluation (i.e., GLEE is not used).

For the ClueWeb09 data (\Fig{clueweb-results}), the serial code trains on a single block (1M
examples) using 9 different ensemble sizes: 100, 250, 500, 750, 1000,
1250, 1500, 1750, 2000. The accuracy ranges from 91.8\% (100 ensemble members) up to 93.8\% (2000 members). The
training time ranges from 12min to 5hr.  COMET trains on 200 blocks
(200M examples), varying across 13 different values for the local
ensemble size: 1, 5, 10, 25, 50, 75, 100, 200, 300, 400, 500,
750, 1000. The total ensemble size is 200 times the local ensemble
size; thus, the largest total ensemble has 200K members.  The accuracy
ranges from 89.5\% (corresponding to a local ensemble size of 1 and a
total ensemble size of 200) to 94.2\% (corresponding to a local
ensemble size of 1000 and a total ensemble size of 200K) with time varying
from less than 1min to 3hr, respectively.  As a point of comparison,
the distributed COMET model achieves an accuracy of 93.8\% (the same as the
best serial model) in only 60min, corresponding to a total ensemble
size of 60K (300 trees per block). Thus, we achieve a 5X speed-up in
training time using 200X more data without sacrificing any accuracy.

On the eBird data (\Fig{ebird-results}), serial IVoting trains from a
single block containing 70K examples and uses the same 9 ensemble
sizes as for the ClueWeb09 data.  The accuracy ranges from 76.4\%
(for the smallest ensemble) up to 77.6\% (for the largest ensemble
time), and training time ranges from 1--20min.  COMET trains on 14
blocks (1M examples), varying across 8 different values for the local
ensemble size: 25, 50, 75, 100, 150, 250, 500, 750. The total
ensemble size is 14 times the local ensemble size; thus, the largest
total ensemble has 10,500 members.  The accuracy ranges from 77.7\%
(better than the best serial accuracy) to 78.9\% with time ranging
from 2min to 9min.  The best accuracy achieved by the serial
version is 77.5\% with a total ensemble size of 2000 and a training
time of 21min; the distributed version improves on this with an
accuracy of 77.8\% for a total ensemble size of only 350 (local size
of 25) and a training time of 2min. Thus, we see a 10X speed-up in
training time while using 14X more data.

\begin{figure*}
  \centering
    \subfloat[ClueWeb Evaluation Savings from GLEE]{\label{fig:glee-clueweb-votes}\includegraphics[scale=0.7]{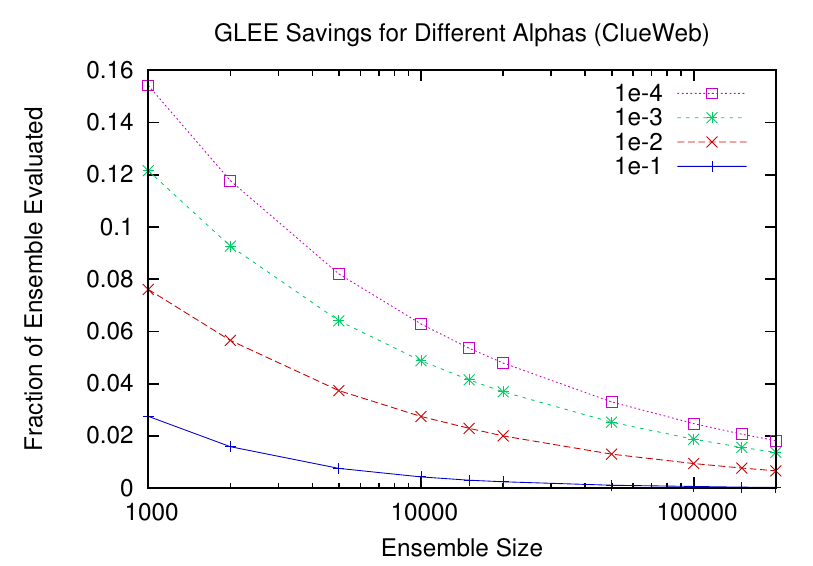}}~~
    \subfloat[ClueWeb Relative Error from GLEE]{\label{fig:glee-clueweb-err}\includegraphics[scale=0.7]{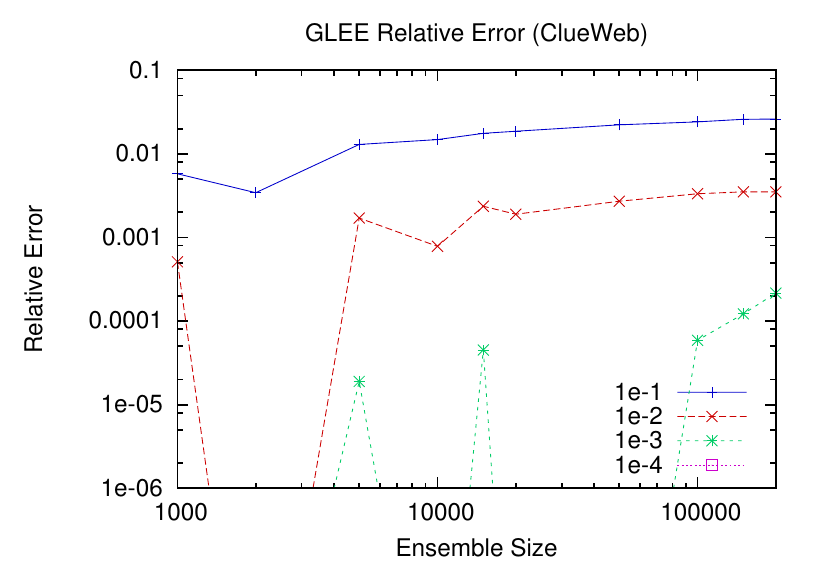}}~~
    \subfloat[Log-log histogram of GLEE early stopping points for an ensemble of size 10000 and
    $\alpha = 0.01$ on ClueWeb data.]{\label{fig:clueweb-evr-stopping-point}\includegraphics[scale=0.7]{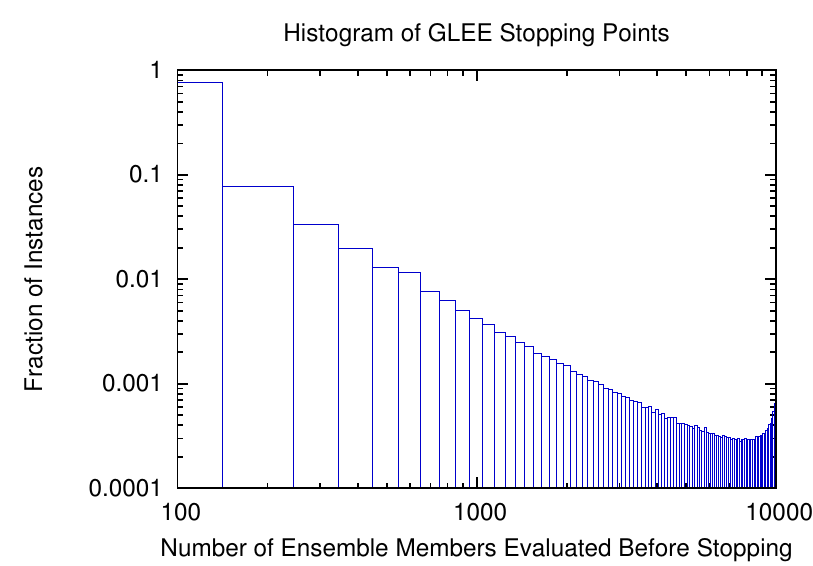}}
  \\
  \vspace{-1ex}
    \subfloat[eBird Evaluation Savings from GLEE]{\label{fig:glee-ebird-votes}\includegraphics[scale=0.7]{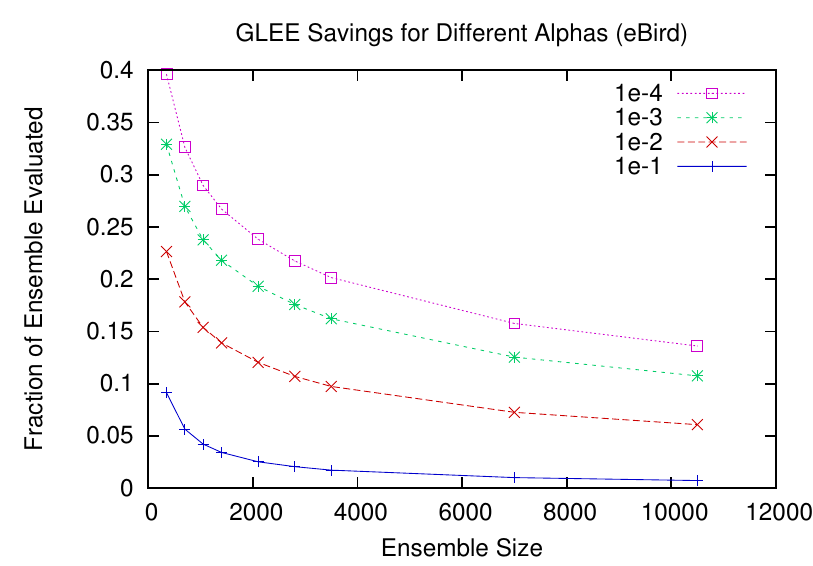}}~~
    \subfloat[eBird Relative Error from GLEE]{\label{fig:glee-ebird-err}\includegraphics[scale=0.7]{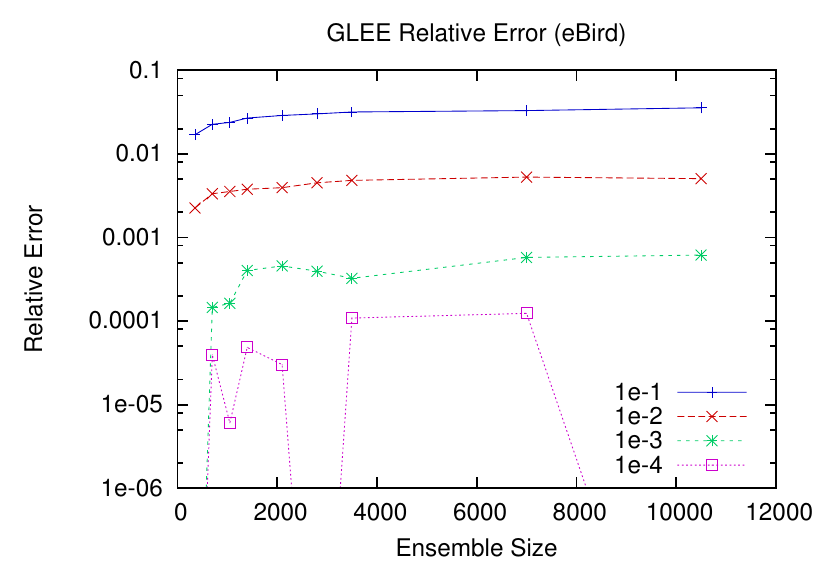}}
    \hspace*{20em}
\shortonly{
  \caption{For both the ClueWeb and eBird tasks, GLEE significantly
    reduces the average cost of ensemble evaluation while maintaining
    relative error below prescribed bounds.  The line graphs plot GLEE
    (called G1-FPC in Fig.~\ref{fig:simulation}) using different
    $\alpha$ values ($1 - $ confidence level).
    Subfigures~\subref{fig:glee-clueweb-votes}
    and~\subref{fig:glee-ebird-votes} show that the fraction of the
    ensemble evaluated decreases as ensemble size grows.  In
    subfigures~\subref{fig:glee-clueweb-err} and
    \subref{fig:glee-ebird-err}, the relative error introduced by GLEE
    is at most $\alpha$, the specified error rate of the
    approximation.  The $\alpha=10^{-4}$ line in
    \subref{fig:glee-clueweb-err} always had less than 1e-6 relative
    error and thus does not appear on the graph.
    Subfigure~\subref{fig:clueweb-evr-stopping-point} shows the
    frequency of the number of votes required. A few difficult cases
    require the full number of votes and are still uncertain.}
}
\longonly{
  \caption{ Relative votes and error for various ensemble sizes and
    values of $\alpha$ using one-tailed Gaussian with
    finite-population correction (G1-FPC) with a minimum of 15 votes.
    Values are calculated relative to using the entire ensemble.}
}
  \label{fig:glee-perf}
  \vspace{-3ex}
\end{figure*}

Figures~\ref{fig:clueweb-data} and~\ref{fig:ebird-data}
 vary the number of data blocks
used in the training. For ClueWeb, all parameters are the same as
above except for the following. The number of blocks is varied from 1
to 200 (with 1M examples per block), and the local ensemble size is
varied from 1 to 1000.  We clearly see a flattening out as the number
of blocks increases, essentially flat-lining at 40. Likewise, the gain
for increasing the ensemble size becomes small (invisible in this
graph) for a local ensemble size of more than 250. For eBird, all
parameters are the same as above except that we fix the local ensemble
size at 200 and vary the number of blocks between 1 and 14. The
accuracy increases almost monotonically with the number of blocks
used.

Figures~\ref{fig:clueweb-results} and~\ref{fig:ebird-results} also show the
performance of COMET using bagging instead of IVoting at local nodes
(COMET-B; tree construction is still randomized).  While COMET-B is
faster than COMET, it is less accurate than both serial and
distributed IVoting (i.e., standard COMET).

The second set of experiments measures the evaluation savings and
relative error incurred by using GLEE for ensembles of different sizes
on the ClueWeb and eBird data
(Figure~\ref{fig:glee-perf}).\footnote{To avoid confounding effects,
  lazy committee evaluation (section~\ref{sec:committee-eval}) is
  \emph{not} used here.  The 200K full-size ClueWeb trees exceeded a
  single node's memory, so an ensemble of 200K trees trained to
  maximum depth of 6 was used for Figure~\ref{fig:glee-perf} instead.}
As expected, the results show that decreasing $\alpha$ increases the
average number of votes (Figures \ref{fig:glee-clueweb-votes},
\ref{fig:glee-ebird-votes}) and decreases the relative error for any
size ensemble (Figures \ref{fig:glee-clueweb-err},
\ref{fig:glee-ebird-err}).  For all ensemble sizes and $\alpha$ values
evaluated, using GLEE provides a significant speed-up over evaluating
the entire ensemble.  This speed-up increases with ensemble size, even
for small values of $\alpha$.  For the ClueWeb data (top row),
relative error is less than 1\% for $\alpha = 0.01$. For an ensemble
of size 1K, fewer than 8\% of the ensemble needs to be evaluated, on
average, and for an ensemble of size 100K, that drops to less than
1\%.  Similar results hold for eBird (bottom row).  Thus, the cost of
evaluating a large ensemble can be largely mitigated via GLEE.
  
Finally, \Fig{clueweb-evr-stopping-point} shows a histogram of the number of evaluations needed by GLEE with $\alpha = 0.01$ on a log-log scale for ClueWeb, providing insight into why the stopping method works --- the vast majority of instances require evaluating only a small proportion of the ensemble.  E.g., 75\% of instances require 100 or fewer base model evaluations.

\section{Related Work} \label{sec:related}

\subsection{Distributed Ensembles}

Ensemble learning has long been used for large-scale distributed
machine learning.  Instead of converting a learning algorithm to be
natively parallel, run the (unchanged) algorithm multiple times, in
parallel, on each data partition
\cite{Chan95:ArbiterCombinerTrees,Domingos96:rule_partitions,Fan02:DynamicScheduling,Chawla04:bites}.
An aggregation strategy combines the set of learned models into an
ensemble that is usually as accurate, if not more accurate, than a
single model trained from all data would have been.  For example, Chan
and Stolfo \cite{Chan95:ArbiterCombinerTrees} study different ways to
aggregate decision tree classifiers trained from disjoint partitions.
They find that voting the trees in an ensemble is sufficient if the
partition size is big enough to produce accurate trees.  They propose
arbiter trees to intelligently combine and boost weaker trees to form 
accurate ensembles in spite of small partitions.  Domingos
\cite{Domingos96:rule_partitions} similarly learns sets of decision
rules from partitioned data, but combines them using a simpler
weighted vote.  Yan et al.~\cite{Yan2009:multimedia} train many
randomized support vector machines with a MapReduce job; a second job runs forward
stepwise selection to choose a subset with good performance.  The
final ensemble aggregates predictions through a simple vote.  In this
work we use simple voting as our aggregation strategy because our data
partitions are relatively large.

Our distributed learning strategy is inspired by Chawla et al.'s work
on distributed IVoting~\cite{Chawla04:bites}.  They empirically
compare IVoting applied to all training data to distributed IVoting
(DIVoting) in which IVoting is run independently on disjoint data
partitions to create sub-ensembles that are merged to make the final
ensemble.
Their results show that DIVoting achieves comparable classification
accuracy to (serial) IVoting with a faster running time, and better
accuracy than distributed bagging that used the same sample sizes.
Compared to DIVoting, COMET benefits from using MapReduce instead of
MPI (for an easier implementation, scaling to data larger than the memory of all nodes, and ability to handle node failures) and incorporates lazy ensemble
evaluation for efficient predictions from large ensembles.  Lazy
evaluation is particularly important when learning from large data
sets with many data partitions.  The work of Wu et al.~\cite{Wu09} is
also closely related to ours.  They also train a decision tree
ensemble using MapReduce in a single pass, but only train one decision
tree per partition, do not use lazy ensemble evaluation, and evaluate
the ensemble on a single small data set with only 699 records.

Like COMET and DIVoting, distributed
boosting~\cite{Yu01:parallel_ensembles, Lazaravic02:parallel_boost}
trains local ensembles from disjoint data partitions and combines them
in a global ensemble.  Worker nodes train boosted trees from local
data but need to share learned models with each other every iteration
to update the sampling weights.  The resulting ensembles are at least
as accurate as boosted ensembles trained serially and can be trained
much faster~\cite{Lazaravic02:parallel_boost}.  Svore and
Burges~\cite{Svore11:parallel_boosting} experiment with a variant of
distributed boosting in which only one tree is selected to add to the
ensemble at each iteration.  As a result the boosted ensemble grows
slowly but is not as accurate as serial boosting.  Chawla et
al.~\cite{Chawla04:bites} showed that DIVoting gives similar accuracy
as distributed boosting without the communication overhead of sharing
models.  

The BagBoo algorithm~\cite{Pavlov10:BagBoo} creates a bagged ensemble
of boosted trees with each boosted sub-ensemble trained independently
from data subsamples.  Like COMET, BagBoo is implemented on MapReduce
and creates mega-ensembles when applied to massive datasets (e.g.,
1.125 million trees).  The ensembles are at least as accurate as
boosted ensembles.  Unlike COMET, sub-ensembles are small (10--20
models) to mitigate the risk of boosting overfitting, and the
non-uniform weights of trees in the ensemble precludes lazy ensemble
evaluation.  Because each sub-ensemble is trained from a sub-sample, a
data point can appear in multiple bags (unlike COMET's partitions); it
is unclear from the algorithm description what communication cost this
incurs.  Since IVoting and AdaBoost yield similar
accuracies~\cite{Breiman99:PastingVotes}, we expect that COMET and
BagBoo would as well.

A different strategy is to distribute the computation for building a
single tree; this subroutine is used to build an ensemble in which
every model benefits from all training data.  This approach involves
multiple iterations of compute nodes calculating and sending split
statistics for their local data to a controller node that chooses the
best split.  Most such algorithms use MPI because of the frequent
communications~\cite{Ye2009:DistributedBoosting,
  Tyree11:parallel_boosting}.  One exception is
PLANET~\cite{Panda2009:PLANET} which constructs decision trees from
all data via multiple MapReduce passes.
PLANET constructs individual trees by treating the construction of each node in the tree as a task involving the partially constructed tree.
The mappers look at the examples that fall into the unexpanded tree node, collect sufficient statistics about each feature and potential split for the node, and send this information to the reducers.
The reducers evaluate the best split point for each feature on a node.
The controller chooses the final split point for the node based on the reducer output.
Because many MapReduce jobs will be involved in building a single tree, PLANET includes many optimizations to reduce overhead, including 1) batching together node construction tasks so that each level in the tree is a single job; 2) finishing subtrees with a small number of items in-memory in the reducer; and 3) using a custom job control system to reduce job setup and teardown costs.

\longonly{
Finally, Moretti et al.~\cite{Moretti2008:cloud_classifiers} build a framework
to support easy learning of distributed ensembles.  The framework is
like MapReduce but with builtin support for specialized data
partitioning and the ability to specify data as test data (to avoid
storing it in replicated file systems).  They demonstrate that their
framework scales on synthetic datasets as big as 54GB.  
}

\subsection{Lazy Ensemble Evaluation}

Whereas much research has studied removing unnecessary models from an
ensemble (called ensemble pruning) \cite{Tsoumakas09:pruning}, only a few studies have used lazy ensemble evaluation to dynamically speed up
prediction time in proportion to the ease or difficulty of each data
point.  Fan et al.~\cite{Fan02:DynamicScheduling} use a Gaussian
confidence interval to decide if ensemble evaluation can stop early
for a test point.  Their method differs from the one described in
Section~\ref{sec:glee} in that a) ensemble members are always
evaluated from most to least accurate, and b) confidence intervals are
based on where evaluation could have reliably stopped on validation
data.  A fixed ordering is not necessary in our work because the base
models have equal voting weight and similar accuracy; this leads to a simpler Gaussian
lazy ensemble evaluation rule.

Markatopoulou et al.~\cite{Markatopoulou10:LEE}
propose a more complicated runtime ensemble pruning, where the choice
of which base models to evaluate is decided by a meta-model trained to
choose the most reliable models for different regions of the input
data space.  Their method can achieve better accuracy than using the
entire ensemble, but generally will not lead to faster ensemble
predictions.

\section{Conclusion} \label{sec:conclusion}

COMET is a single-pass MapReduce algorithm for learning on large-scale
data.  It builds multiple ensembles on distributed blocks of data and
merges them into a mega-ensemble.  This approach is appropriate when
learning from massive-scale data that is too large to fit on a single
machine.  It compares favorably (in both accuracy and training time)
to learning on a subsample of data using a serial algorithm.  Our
experiments showed that it is important to use a sequential ensemble
method (IVoting in our case) when building the local ensembles to get
the best accuracy.

The combined mega-ensemble can be efficiently evaluated using lazy
ensemble evaluation; depending on the ensemble size, the savings in
evaluation cost can be 100X or better.  Two options are available for
lazy evaluation: our GLEE rule and the Bayesian MLEE
rule~\cite{HernandezLobato09:InstanceBasedPruning}.  GLEE is easy to
implement, is asymptotically faster to compute than MLEE, and provides
the same evaluation savings and approximation quality as MLEE.
If one desires to further speed up evaluation or reduce the model's
storage requirements, ensemble pruning~\cite{Tsoumakas09:pruning}
could be applied to remove extraneous base models, or model
compression~\cite{Bucila06:munge} could be used to compile the
ensemble into an easily deployable neural network.  Ultimately the
appropriateness of sacrificing some small accuracy (and how much
accuracy) for faster evaluations will depend on the application
domain.

In future work, it will be interesting to contrast COMET to
PLANET~\cite{Panda2009:PLANET}, which builds trees using all available
data via multiple MapReduce passes.  As there is no open-source
version of PLANET currently available and this procedure is highly
time-consuming without special modifications to
MapReduce~\cite{Panda2009:PLANET}, we are unable to provide direct
comparisons at this time.  However, we imagine that there will be some
trade-off between accuracy (using all data for every tree) and time
(since COMET uses only a single MapReduce pass).

\longonly{
\begin{figure}[tbhp]
  \centering
    \includegraphics[scale=1]{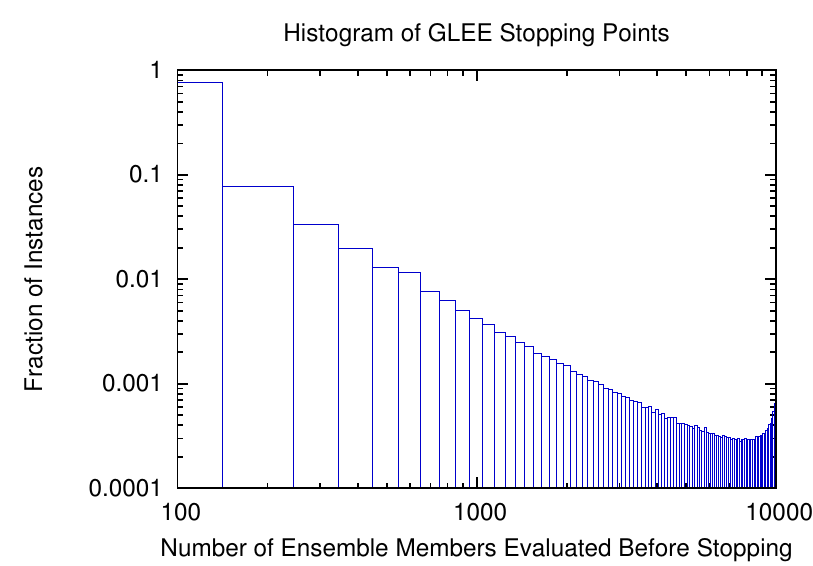}
  \caption{Histogram of stopping point of ensemble evaluation using
    one-tailed Gaussian with finite-population correction (G1-FPC) on
    an ensemble of size 10000 and $\alpha = 0.01$ for the ClueWeb
    data.}
  \label{fig:clueweb-evr-stopping-point}
\end{figure}
}

\section*{Acknowledgments}  
\begin{small}
The authors thank David Gleich, Todd Plantenga, and Greg Bayer for many helpful discussions.  We are particularly grateful to David for suggesting the learning task with the ClueWeb data.

 Sandia National Laboratories is a multi-program laboratory managed and operated by Sandia Corporation, a wholly owned subsidiary of Lockheed Martin Corporation, for the U.S. Department of Energy's National Nuclear Security Administration under contract DE-AC04-94AL85000.
\end{small}


\appendix

\section{Data Preprocessing}
\label{app:preproc}

We performed several preprocessing steps on the eBird data.  First, we
removed records for checklists covering $> 5$ miles since most
attributes are based on a checklist's location and the location
information for checklists covering large distances is less reliable.
Second, count types P34 and P35 are rare in the data, so we grouped
them with the semantically equivalent count types P22 and P23,
respectively, by recoding P34 as P22 and P35 as P23.  Third, we
rederived the attributes \attr{caus\_temp\_avg},
\attr{caus\_temp\_min}, \attr{caus\-\_temp\-\_max}, \attr{caus\_prec},
and \attr{caus\_snow} using the \attr{month} attribute and the
appropriate monthly climate features (e.g., \attr{caus\_temp\_avg01})
to remove thousands of spurious missing values.  (The eBird data
providers plan to correct this processing mistake in future versions.)
Fourth, missing values for categorical attributes were replaced with
the special token `MV'.  Similarly, missing and NA values for
numerical attributes were recoded as -9999, a value outside the
observed range of the data.  With these missing value encodings, the
decision tree learning algorithm can handle records with missing
values as a special case if that leads to better accuracy.  Fifth, we
converted the categorical features (\attr{count\_type}, \attr{bcr},
\attr{bailey\_ecoregion}, \attr{omer\-nik\-\_l3\-\_eco\-region}) into
multiple binary features with one feature per valid feature value
because our decision tree implementation does not yet handle
categorical features.

\end{document}